\documentclass[conference]{IEEEtran}
\IEEEoverridecommandlockouts
\usepackage{cite}
\usepackage{amsmath,amssymb,amsfonts}
\usepackage{algorithmic}
\usepackage{graphicx}
\usepackage{textcomp}
\usepackage{xcolor}
\usepackage{multirow}
\usepackage{stfloats}
\usepackage{url}
\def\BibTeX{{\rm B\kern-.05em{\sc i\kern-.025em b}\kern-.08em
    T\kern-.1667em\lower.7ex\hbox{E}\kern-.125emX}}
\begin{document}

\title{Reconstruction of Perceived Images from fMRI Patterns and Semantic Brain Exploration using Instance-Conditioned GANs\\
\thanks{Funded by AI-REPS grant ANR-18-CE37-0007-01 and ANITI grant ANR-19-PI3A-0004}
}
%\author{\IEEEauthorblockN{Anonymous Authors}}
\makeatletter
\newcommand{\linebreakand}{%
  \end{@IEEEauthorhalign}
  \hfill\mbox{}\par
  \mbox{}\hfill\begin{@IEEEauthorhalign}
}
\makeatother

\author{

\IEEEauthorblockN{Furkan Ozcelik}
\IEEEauthorblockA{\textit{CerCo, CNRS UMR5549} \\
\textit{Universit\'e de Toulouse}\\
%\textit{CNRS}\\
\textit{Toulouse, France} \\
furkan.ozcelik@univ-tlse3.fr}
\and
\IEEEauthorblockN{Bhavin Choksi}
\IEEEauthorblockA{\textit{CerCo, CNRS UMR5549} \\
\textit{Universit\'e de Toulouse}\\
%\textit{CNRS}\\
\textit{Toulouse, France} \\
bhavin.choksi@cnrs.fr}
\and
\IEEEauthorblockN{Milad Mozafari}
\IEEEauthorblockA{\textit{CerCo, CNRS UMR5549} \\
\textit{IRIT, CNRS UMR5505} \\
%\textit{CNRS}\\
\textit{Toulouse, France} \\
milad.mozafari@cnrs.fr}

\linebreakand

\IEEEauthorblockN{Leila Reddy}
\IEEEauthorblockA{\textit{CerCo, CNRS UMR5549 and} \\
\textit{ANITI, Universit\'e de Toulouse}\\
\textit{Toulouse, France} \\
leila.reddy@cnrs.fr}
\and
\IEEEauthorblockN{Rufin VanRullen}
\IEEEauthorblockA{\textit{CerCo, CNRS UMR5549 and} \\
\textit{ANITI, Universit\'e de Toulouse}\\
\textit{Toulouse, France} \\
rufin.vanrullen@cnrs.fr}
}

%\and
%\and
%\and
%\and
%\and

\maketitle

\begin{abstract}
Reconstructing perceived natural images from fMRI signals is one of the most engaging topics of neural decoding research. Prior studies had success in reconstructing either the low-level image features or the semantic/high-level aspects, but rarely both. In this study, we utilized an Instance-Conditioned GAN (IC-GAN) model to reconstruct images from fMRI patterns with both accurate semantic attributes and preserved low-level details. The IC-GAN model takes as input a 119-dim noise vector and a 2048-dim instance feature vector extracted from a target image via a self-supervised learning model (SwAV ResNet-50); these instance features act as a conditioning for IC-GAN image generation, while the noise vector introduces variability between samples. We trained ridge regression models to predict instance features, noise vectors, and dense vectors (the output of the first dense layer of the IC-GAN generator) of stimuli from corresponding fMRI patterns. Then, we used the IC-GAN generator to reconstruct novel test images based on these fMRI-predicted variables. The generated images presented state-of-the-art results in terms of capturing the semantic attributes of the original test images while remaining relatively faithful to low-level image details. Finally, we use the learned regression model and the IC-GAN generator to systematically explore and visualize the semantic features that maximally drive each of several regions-of-interest in the human brain. 
%(ROI, explainability, perturbation, nearest neighbor
%(...) analyses. These analyses provide a more detailed understanding of the semantic information included in the fMRI patterns.
\end{abstract}


\section{Introduction}

Understanding the brain and cognition has always been one of the fundamental research areas of science. One of the ways researchers approach this task is by establishing neural encoding and decoding methods. New ways to decode information from brain signals have emerged with recent developments in modeling and computation.
 
In vision research, many studies have used statistical methods and machine learning to decode specific information like position~\cite{thirion2006inverse} or orientation~\cite{kamitani2005decoding,haynes2005predicting}, to classify image categories~\cite{haxby2001distributed,cox2003functional}, to retrieve the closest images from a candidate set~\cite{kay2008identifying}, or even to reconstruct images with low-complexity like basic shapes and structures~\cite{miyawaki2008visual}. 

With the emergence of deep learning, and in particular advanced deep generative models, reconstructing more complex images like handwritten digits~\cite{schoenmakers2013linear}, faces~\cite{vanrullen2019reconstructing}, and natural scenes~\cite{shen2019deep} has become possible. These deep generative models include variational auto-encoders (VAEs), generative adversarial networks (GANs), and many variants and hybrids of both. Although many studies have used these models, they typically managed to reconstruct either low-level or high-level features of the images, but rarely both at the same time. 

Here, we propose a method to reconstruct natural images from fMRI activation patterns with both accurate semantic attributes and relatively preserved low-level details, using an Instance-Conditioned GAN (IC-GAN) -- a recent generative model~\cite{casanova2021instanceconditioned} inspired by the success of self-supervised feature learning~\cite{jaiswal2021survey}. In our framework, we first extract latent representations for a set of training images (see Figure~\ref{fig:latent_extraction}): high-level attributes of the images, called ``instance features'' in IC-GAN, are computed with a single forward pass through the SwAV ResNet-50 model; lower-level aspects of the image (e.g. reflecting the size, position or orientation of an object, details of the background, etc.) are obtained via a two-stage optimization of the IC-GAN ``noise'' and ``dense'' latent vectors (inspired by the method of Pividori et al.~\cite{pividori2019exploiting}). Next, we train three ridge regression models to predict these latent image representations from the corresponding fMRI patterns, recorded while human subjects viewed the same training images (Figure~\ref{fig:decoding_reconstruction}, Step 1). Finally, for each image in the test set, we predict the instance feature, noise vector, and dense vector from fMRI data (using the previously learned regression models), and then reconstruct an estimate of the image using IC-GAN's generator (Figure~\ref{fig:decoding_reconstruction}, Step 2 and 3). The code of this paper can be found in the official GitHub repository\footnote{\texttt{\url{https://github.com/ozcelikfu/IC-GAN_fMRI_Reconstruction}}}.
    
Our method establishes a new state-of-the-art performance for capturing the semantic attributes of the images, while preserving a reasonable amount of low-level details. We present both qualitative and quantitative results, and a comparison with previous methods to support our claims. We also take advantage of our brain-based image reconstruction system to explore and visualize the semantic image attributes encoded in various brain regions-of-interest (ROIs), and discuss how these findings align with neuroscientific evidence. 

\section{Previous Works}

Many methods have been proposed in the literature for reconstructing natural images from fMRI patterns using deep learning models. Shen et al.~\cite{shen2019deep} optimized input images with a deep generator network using the loss provided by fMRI decoded CNN features. Seeliger et al.~\cite{seeliger2018generative} trained a linear regression between fMRI signals and DCGAN's latent space using a feature loss from a CNN model and a pixel-space MSE loss. In addition to supervised training with \{fMRI, stimulus\} pairs, Beliy et al.~\cite{beliy2019voxels} used a consistency loss for unsupervised training with test fMRI data (without corresponding stimuli) and additional image data. Later, Gaziv et al.~\cite{gaziv2020self} improved on this method by imposing a perceptual loss on reconstructed images, resulting in sharper reconstructions. Mozafari et al.~\cite{mozafari2020reconstructing} proposed the first semantically-oriented reconstruction model using BigBiGAN~\cite{donahue2019large}. Finally, Ren et al.~\cite{ren2021reconstructing} developed a dual VAE-GAN model that uses a three-stage learning strategy incorporating adversarial learning and knowledge distillation. We qualitatively and quantitatively compare the results of these studies with our proposed method.

\section{Materials and Methods}

\begin{figure}
    \centering
    \includegraphics[width=0.5\textwidth]{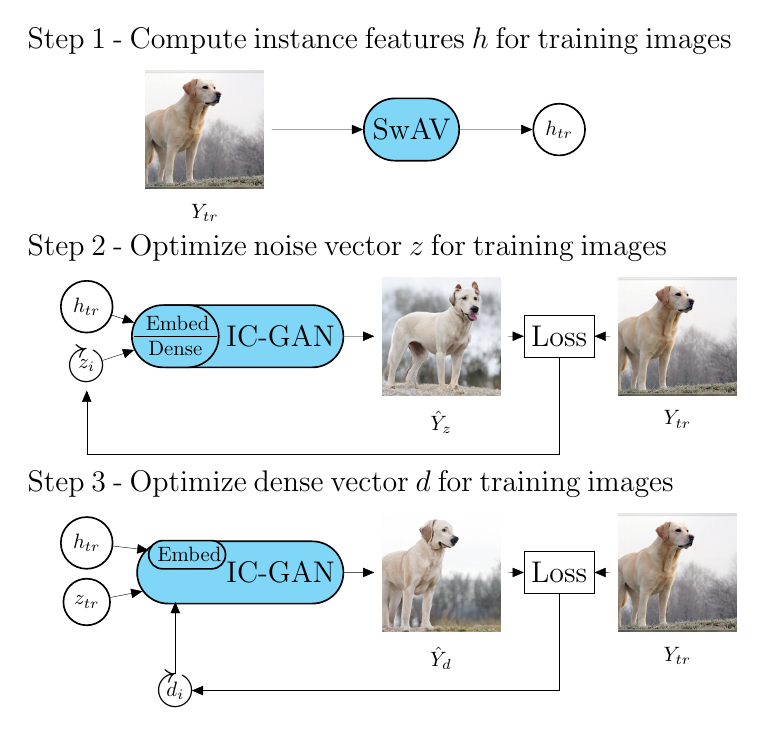}
    \caption{Extraction of the latent variables ($h_{tr}$, $z_{tr}$ and $d_{tr}$) for each training image ($Y_{tr}$). Step 1: Instance features of training images ($h_{tr}$) are extracted using SwAV ResNet-50. This 2048-dim instance feature vector ($h_{tr}$) captures the semantic attributes of the image. Step 2: In addition to the instance feature vector, the IC-GAN also requires a noise vector ($z_{i}$) as input, which encodes lower-level properties of the image (e.g., pose, orientation, background etc.). While providing $h_{tr}$ obtained from Step 1 to the IC-GAN's generator, we optimize the noise vector ($z_{i}$) to generate the closest image ($\hat{Y}_{z}$) to the groundtruth image ($Y_{tr}$). The resulting optimized noise vector is $z_{tr}$. Step 3: To further improve image reconstruction so as to better match the more detailed spatial structure of the training image, we apply another optimization stage, in which we optimize the dense layer vectors of IC-GAN itself. To achieve, this, we pass the first $17$ dimensions of $z_{tr}$ to the dense layer of the IC-GAN's generator and obtain initial dense vectors ($d_{0}$). While keeping both $h_{tr}$ and the remaining $102$ dimensions of $z_{tr}$ fixed, we optimize the dense vector $d_{i}$ to generate the closest image ($\hat{Y}_{d}$) to the groundtruth image ($Y_{tr}$). $d_{tr}$ is the resulting optimized dense vector.}
    \label{fig:latent_extraction}
\end{figure}

\subsection{Instance-Conditioned GAN}
We utilized an Instance-Conditioned GAN (IC-GAN) model, pretrained for natural image generation on the ImageNet dataset~\cite{deng2009imagenet}. IC-GAN can be considered as a generic framework rather than a single model, because it can be applied to different GAN backbones, e.g. StyleGAN~\cite{karras2020analyzing} or BigGAN~\cite{brock2018large}. In the usual conditional GAN setting~\cite{mirza2014conditional}, class labels are provided along with noise vectors sampled from a normal distribution to generate images. Images belonging to that specific class are labeled as ``real'', and generated images from the generator are labeled as ``fake''. Both the generator and discriminator are trained with these images and labels in an adversarial learning framework. In the instance-conditional setting, instead of giving a class label, instance features that capture the semantic attributes of a given image are extracted (via a pre-trained feature extractor) and provided to the generator as conditioning, alongside a sampled noise vector. For training, IC-GAN selects $k$ images in the neighborhood of the conditioning image (according to the feature extractor); these images are labeled as real, while generated images are considered as fake images to train both the generator and discriminator.
         
For the instance feature extraction, IC-GAN models use the SwAV (Swapping Assignments between Views) architecture~\cite{caron2020unsupervised} with a ResNet-50~\cite{he2016deep} backbone. SwAV is a self-supervised learning model which means that it does not require handcrafted labels from humans. Similar to contrastive learning methods~\cite{jaiswal2021survey}, SwAV minimizes the distance in feature space between representations of two transformed images (coming from the same original image).

It is possible to train the IC-GAN framework with different feature extractors, as long as they provide rich feature representations. However, using features from self-supervised learning models (e.g., SwAV) is better suited to the problem of neural decoding and natural image reconstruction. Indeed, many recent studies show that representations gathered from self-supervised learning models present more similarity to brain representations than other learning methods~\cite{konkle2022self,zhuang2021unsupervised}.

The specific IC-GAN model we used here relies on a BigGAN~\cite{brock2018large} architecture with 7 layers. It generates $256\times256\times3$ images from a $2048$-dim (dimensional) instance feature vector extracted from SwAV ResNet-50 and a $119$-dim noise vector sampled from a normal distribution. The $2048$-dim instance features are given to an embedding layer and thus reduced to $512$-dim embedded vectors. The $119$-dim noise vector, which encodes lower-level properties of the image (e.g. pose, size, orientation of the object), is split into seven hierarchical levels, each with $17$ dimensions. The first $17$-dim level is directly given to the first dense layer of the IC-GAN generator. The remaining six hierarchical levels are concatenated with the embedded instance vector to be fed to the generator in each of the six BigGAN residual blocks. 

Overall, the purpose of IC-GAN is to generate, from one conditioning image, new and diverse image instances that share semantic attributes (as captured by SwAV instance features), but differ in low-level properties (e.g. object position, size, orientation, background details). The diversity of low-level properties is determined by randomly sampled ``noise'' vectors (and by the ``dense'' vectors directly derived from them). However, for the purpose of fMRI-based image reconstruction, both high-level and low-level properties must be specified. Therefore, rather than randomly sampling noise vectors, we computed a specific noise vector (and the associated dense vector) for each training image in the dataset, as detailed below.

\subsection{Extracting Latent Variables from Training Stimuli}

We illustrate the computation of latent variables in Figure~\ref{fig:latent_extraction}. We first extracted a $2048$-dim instance feature vector for each training image in our dataset (see dataset details below) by presenting it to a SwAV ResNet-50 feature extractor. We then provided these instance features to the IC-GAN generator, and optimized the $119$-dim noise vector for the same image using the covariance matrix adaptation evolution strategy (CMA-ES)~\cite{hansen2001completely}. We used this method because we empirically observed that global optimization strategies worked better than local optimization strategies (like gradient-based methods) for the noise vector. The loss function for this optimization was the distance between the generated image and the original training image in Layer$-4$ of SwAV ResNet-50; this representation level, hierarchically lower than the instance feature level, encodes more spatially structured information.

Finally, to further match the more detailed spatial structure of the original image, we applied one more optimization stage. Inspired by the two-stage inversion method of Pividori et al.~\cite{pividori2019exploiting}, we provided the first $17$ dimensions of the previously optimized noise vector to the first dense layer of the IC-GAN, resulting in a $1536\times4\times4$-dim dense vector. While the instance features and the remaining $102$ dimensions of the noise vector were kept fixed, we optimized these dense vectors with the RMSProp optimizer. For this second-stage optimization, the previous loss (SwAV ResNet-50 Layer$-4$ feature distance) was combined with a Learned Perceptual Image Patch Similarity (LPIPS)~\cite{zhang2018unreasonable} loss gathered from a pretrained VGG16 model~\cite{simonyan2014very} and a pixel (MSE) loss from $64\times64$ resized images.

\begin{figure}[htbp]
    \centering
    \includegraphics[width=0.5\textwidth]{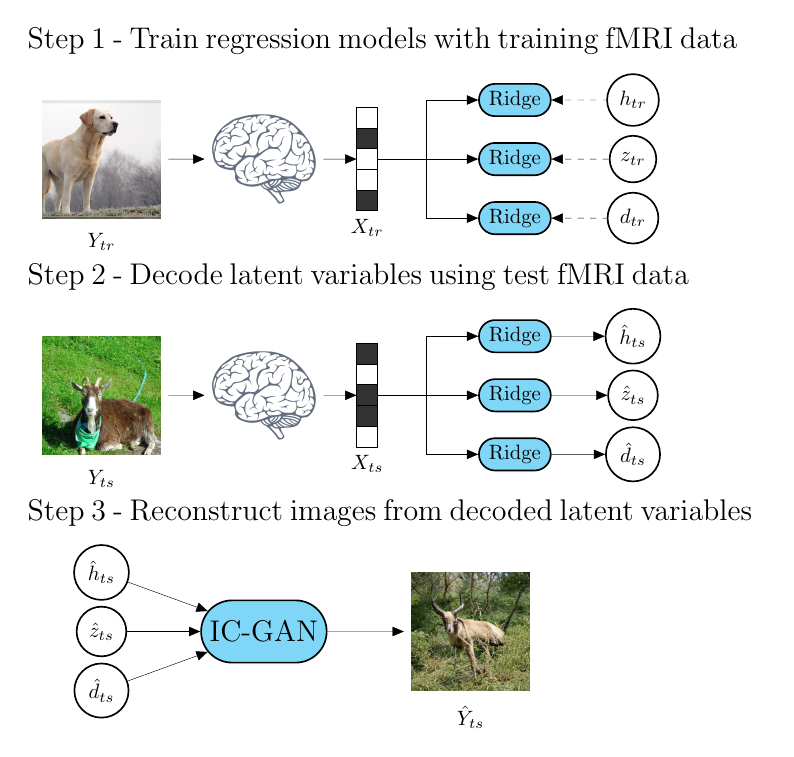}
    \caption{Decoding latent variables from fMRI patterns and reconstructing images from decoded variables. Step 1: Having obtained the instance features ($h_{tr}$), noise vectors ($z_{tr}$) and dense vectors ($d_{tr}$) of training images ($Y_{tr}$) as described in Figure~\ref{fig:latent_extraction}, we train three ridge regression models to map fMRI patterns of the training set ($X_{tr}$) to these latent variables. Step 2: Using these trained regression models, we decode latent variables of the test set ($\hat{h}_{ts}$,$\hat{z}_{ts}$,$\hat{d}_{ts}$) from test fMRI patterns ($X_{ts}$). Step 3: We pass the decoded latent variables to the IC-GAN Generator to obtain reconstructed images ($\hat{Y}_{ts}$) }
    \label{fig:decoding_reconstruction}
\end{figure}

\subsection{Generic Object Decoding Dataset}

In this study, we used previously published fMRI recordings of five human subjects presented with images from the ImageNet dataset~\cite{horikawa2017generic}. The dataset contains training and testing image perception sessions where subjects looked at 1200 training samples drawn from 150 categories (8 samples each) and 50 testing samples chosen from 50 categories (1 sample each), respectively. Training and testing categories were chosen independently and were non-overlapping. Each training image was presented only once, while testing images were repeated 35 times during the whole experiment. All fMRI runs followed a similar design: fixation (33s), 50 image presentations (9s per image flashing at 2Hz), fixation (6s). Moreover, subjects were also asked to perform a one-back task by pressing a button whenever the same image was presented two times in a row (five such events occurred per run).

The fMRI data were pre-processed for each subject by three-dimensional motion correction followed by coregistration to the high-resolution anatomical image. Then, the brain representation of each image was calculated by averaging the percent signal change values of each voxel over the 9-s presentation window. Additionally, the dataset provides functional regions of interest (ROIs) that cover the entire visual cortex, including V1-V4, the fusiform face area (FFA), parahippocampal place area (PPA), and lateral occipital complex (LOC). The pre-processed data is available to download at brainliner.jp\footnote{\texttt{\url{http://brainliner.jp/data/brainliner/Generic_Object_Decoding}}}.

\subsection{fMRI Decoding and Image Reconstruction}
 Details of fMRI decoding and image reconstruction are depicted in Figure~\ref{fig:decoding_reconstruction}. The procedure involves two separate stages for training and testing the brain decoding system of each subject. 
 
 First, we trained three separate ridge regression models to predict the latent variables (instance features; noise vectors; dense vectors) for each of the 1200 training images based on the corresponding fMRI patterns. Since both the fMRI data and the latent variables are high-dimensional, we applied $L_2$ regularization on the regression weights during training.

At test time, we averaged the 35 repetitions of fMRI signals corresponding to each test stimulus. Next, we used the previously trained regression models to predict the instance features, noise vectors, and dense vectors from these averaged fMRI signals. Finally, we used these predicted latent variables to generate image reconstructions using the IC-GAN generator.

\section{Results and Analyses}

\subsection{Image Reconstruction Results}

\begin{figure*}
    \centering
    \includegraphics[width=0.55\textwidth]{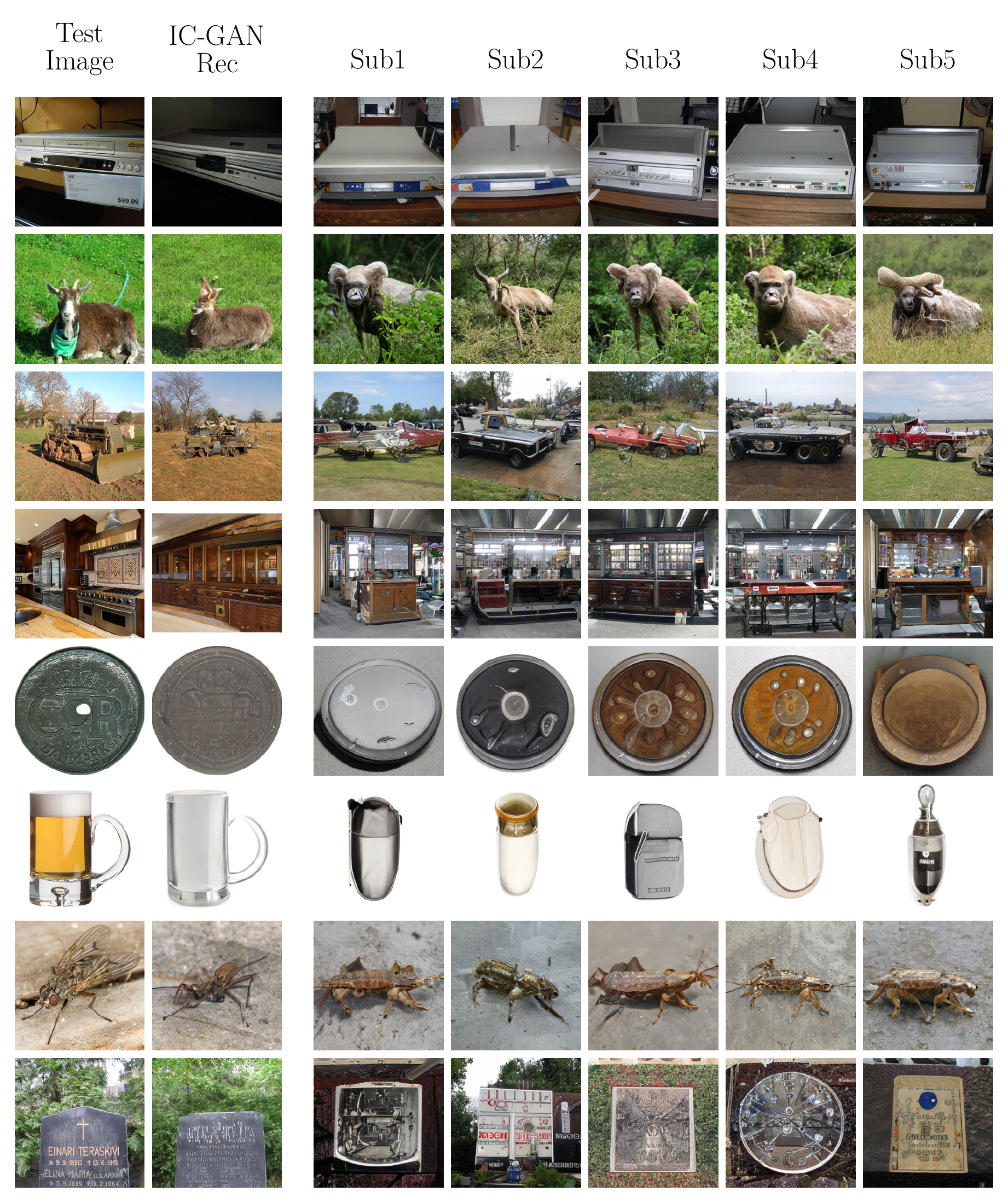}
    \caption{fMRI Reconstructions by the IC-GAN model for all subjects. The first column is the groundtruth test image, whereas the second column is the reconstructed image by IC-GAN using true extracted latent variables. The following five columns demonstrate the equivalent reconstructions using fMRI-decoded latent variables for each subject. fMRI reconstructions are generally consistent with the groundtruth images in terms of semantic attributes, while they preserve the low-level details to a certain degree.}
    \label{fig:multisubject_reconstruction}
\end{figure*}

%\begin{figure*}
%  \hspace{0.025\textwidth}
%  \begin{minipage}[c]{0.6\textwidth}
%    \includegraphics[width=1\textwidth]{figures/MultiSubjectRecs.pdf}
%  \end{minipage}%\hfill
%  \hspace{0.025\textwidth}
%  \begin{minipage}[b]{0.3\textwidth}
  
%    \caption{
%		fMRI Reconstructions by the IC-GAN model for all subjects. The first column is the groundtruth test image, whereas the second column is the reconstructed image by IC-GAN using true extracted latent variables. The following five columns demonstrate the equivalent reconstructions using fMRI-decoded latent variables for each subject. fMRI reconstructions are generally consistent with the groundtruth images in terms of semantic attributes, while they preserve the low-level details to a certain degree.
%    } \label{fig:multisubject_reconstruction}
%  \end{minipage}
%\end{figure*}

\begin{figure*}
    \centering
    \includegraphics[width=0.65\textwidth]{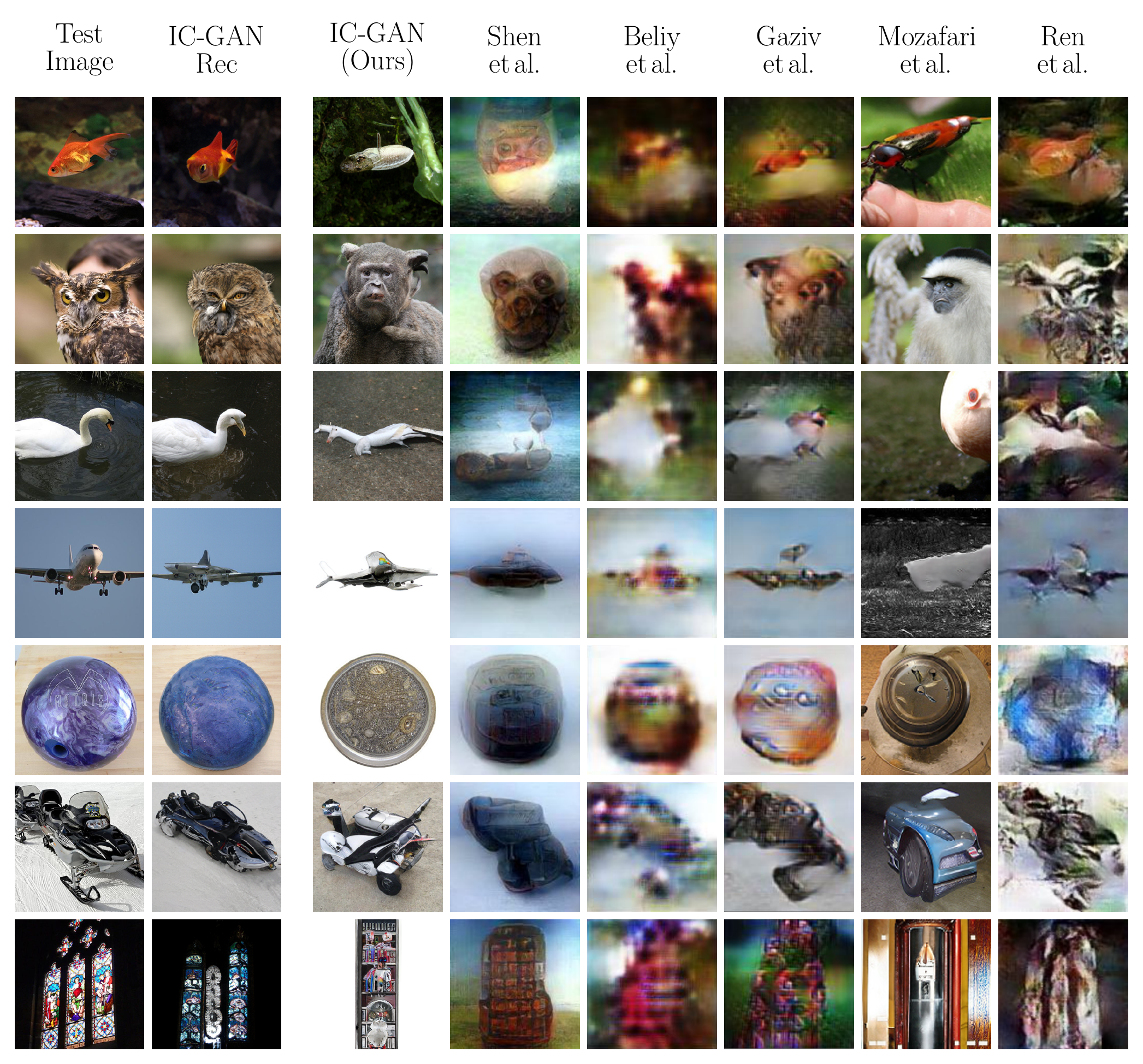}
    \caption{Comparison of fMRI reconstructions for several methods. The first column is the groundtruth image, the second column is the reconstructed image with IC-GAN using true extracted latent variables. Columns three to eight present fMRI reconstructions from IC-GAN (Ours), Shen et al.~\cite{shen2019deep}, Beliy et al.~\cite{beliy2019voxels}, Gaziv et al.~\cite{gaziv2020self}, Mozafari et al.~\cite{mozafari2020reconstructing}, and Ren et al.~\cite{ren2021reconstructing}, respectively. fMRI reconstructions by the IC-GAN method demonstrate more naturalistic-looking images with accurate semantic attributes, while preserving some low-level details (e.g. object position, size or orientation).}
    \label{fig:method_comparison}
\end{figure*}

Examples of image reconstructions produced by our method are displayed in Figure~\ref{fig:multisubject_reconstruction}. First of all, it is important to examine IC-GAN reconstructions (second column) based on the optimized ``ground-truth'' latent vectors (derived as detailed in Figure~\ref{fig:latent_extraction}): we can see that IC-GAN can successfully reconstruct the semantic attributes of the test images; however, it often misses some visual details, like parts of the vehicle (third row), liquid in the glass (fifth row), or the precise text in the gravestone (eighth row). These reconstructions help us understand how the IC-GAN generator would behave if we perfectly decoded latent variables from fMRI patterns, i.e. they serve as an upper bound on the expected reconstruction quality. 

When we inspect actual fMRI reconstructions for the five subjects (third to seventh columns), our first observation is that reconstructions look like natural images. Furthermore, they are consistent across subjects. Again, these reconstructions capture some of the semantic attributes, while also missing specific aspects of the test images. For example, the system generates images of horned animals for the goat image (second row), but their species are not clearly identifiable. For the token image (fifth row), round objects are reconstructed, but not with the right texture. For the gravestone image (last row), similar square-shaped objects with text and symbols are generated, but most of them would not qualify as a gravestone. Overall, our method appears to reconstruct semantic attributes with slight but significant variations in details.

How does it compare to previously proposed methods? In Figure~\ref{fig:method_comparison}, we present image reconstructions using alternative methods proposed in five other studies, together with our results for comparison\footnote{We selected these seven images because it was the only common set of reconstruction exemplars presented across all of the considered studies.}. From these reconstructions, we can see that many methods capture low-level details rather than high-level ones; as a result, many of the reconstructions do not look natural. A notable exception is the study of Mozafari et al., based on the BigBiGAN architecture~\cite{mozafari2020reconstructing}, in which reconstructions often capture high-level properties and are more naturalistic. Even this method, however, does not correctly reconstruct semantic details for some of the images; furthermore, it misses many of the low-level details. Among the other studies, Ren et al.~\cite{ren2021reconstructing} succeed in reconstructing colors and textures better than other methods, while Gaziv et al.~\cite{gaziv2020self} give sharper object edges. Our method generates realistic-looking image reconstructions with appropriate semantic features, while preserving the low-level aspects to a certain degree.

\begin{table}[h]
    \centering
    \caption{Quantitative comparison of image reconstructions. For each measure, the best value is in bold. (For Pix-Comp/SSIM, higher is better; for Inception/CLIP distance, lower is better)}
    \label{tab:method_cmp}
    \begin{tabular}{|l|c|c|c|c|}
        \hline
        \multirow{3}{*}{Method} & \multicolumn{4}{c|}{Similarity Measure}\\
        \cline{2-5}
        ~ & \multicolumn{2}{c|}{Low-Level} & \multicolumn{2}{c|}{High-Level}\\
        \cline{2-5}
        ~ & Pix-Comp $\uparrow$ & SSIM $\uparrow$ & Inception $\downarrow$ & CLIP $\downarrow$\\
        \hline
        \hline
         Shen et al.~\cite{shen2019deep} & $79.7\%$ & $0.582$ & $0.829$ & $0.358$\\
         Beliy et al.~\cite{beliy2019voxels} & $85.3\%$ & $0.597$ & $0.865$ & $0.424$\\
         Gaziv et al.~\cite{gaziv2020self} & \boldmath{$91.5\%$} & \boldmath{$0.601$} & $0.841$ & $0.387$\\
         Ren et al.~\cite{ren2021reconstructing} & $87.3\%$ & $0.588$ &  $0.847$ & $0.383$\\
         %Eigen-Image (PCA) & $73.4\%$ & $0.884$ & \\
         Mozafari et al.~\cite{mozafari2020reconstructing} & $54.3\%$ & $0.450$ & $0.818$ & $0.352$\\
         \hline
         IC-GAN (Random) & $64.1\%$ & $0.467$ & $0.761$ & $0.328$\\
         IC-GAN (Noise)  & $66.5\%$ & $0.489$ & $0.744$ & \boldmath{$0.320$}\\
         IC-GAN (Dense)  & $67.2\%$ & $0.491$ & \boldmath{$0.742$} & $0.330$ \\
         \hline
    \end{tabular}
    
\end{table}

These qualitative observations are supported by the quantitative comparison of methods in Table~\ref{tab:method_cmp}, according to both low-level measures of image quality (Pix-Comp, SSIM) and higher-level ``semantic'' measures (Inception or CLIP distance). Pix-Comp is a 2-way comparison of pixel-wise correlation measures computed over the whole test set. We used the results reported by authors in their respective papers, except for Gaziv et al.~\cite{gaziv2020self}, who did not report Pix-Comp: we re-computed it over the reconstructed images provided in their supplementary material. All other metrics (SSIM~\cite{wang2004image}, Inception-V3~\cite{szegedy2016rethinking} distance, and CLIP ViT-B/32~\cite{radford2021learning} distance), were computed over the seven common image reconstructions presented in Figure~\ref{fig:method_comparison}. Our own results are presented for three different versions of IC-GAN decoding, using different combinations of the three brain regression models in Figure~\ref{fig:decoding_reconstruction}, to evaluate the effects of each regressor on performance. First, the IC-GAN (Random) version uses brain-decoded instance features together with randomly sampled noise vectors from a normal distribution. Second, the IC-GAN (Noise) version combines brain-decoded instance features with brain-decoded noise vectors, without using the brain-decoded dense vectors (instead, the output of the first dense layer is used directly). Finally, IC-GAN (Dense) is the complete framework described in Figure~\ref{fig:decoding_reconstruction}, which uses all the brain-decoded latent variables (thus overriding the dense vector with its brain-decoded version). The table indicates that most other methods yield better results than IC-GAN on the low-level measures (Pix-Comp, SSIM), except for Mozafari et al~\cite{mozafari2020reconstructing}; like ours, that study was aimed at matching higher-level ``semantic'' aspects of the input images. Importantly, IC-GAN outperforms the Mozafari et al method for both low-level measures. For the high-level measures (Inception and CLIP Distances), IC-GAN demonstrates state-of-the-art performance, surpassing all methods--including Mozafari et al.--by a significant margin. 

The comparison of the 3 versions of our IC-GAN method reveals that the inclusion of both the brain-decoded noise vector (IC-GAN Noise) and the brain-decoded dense vector (IC-GAN Dense) helps improve the model's ability to capture low-level details. Still, the full model remains inferior to many previous methods in this respect. Regarding high-level semantic attributes, while the full method IC-GAN (Dense) is superior to IC-GAN (Noise) for the Inception distance, the opposite is true for the CLIP distance. This could be because Inception features include more spatially structured information than CLIP features; indeed, the function of dense vectors in our method is precisely to capture the image spatial structure that is less explicitly encoded in the noise vectors. 

\subsection{Semantic Analysis of Visual Encoding in Brain ROIs}

Our brain decoding model, relying on the latent space(s) of the IC-GAN network, can reconstruct the high-level content of perceived images better than all prior methods, while retaining more low-level details than at least some of these methods. From a neuroscience viewpoint, can this brain decoding model also help us understand the neural coding of visual information in the brain? Here, we use our model to explore and directly visualize the types of information that are preferentially represented in various brain regions-of-interest (ROIs). 

The fMRI dataset counts seven distinct ROIs across visual cortex for each subject--in hierarchical order: V1, V2, V3, V4, LOC (Lateral Occipital Complex), FFA (Fusiform Face Area) and PPA (Parahippocampal Place Area). First, we ask whether each region carries more information about high-level latent features--as captured by the model's instance features--or about low-level properties--as captured by the model's dense vector (note that similar results, not shown here, were obtained for the noise vector instead of the dense vector). To answer this question, for each brain voxel we compared the $L_1$ norm of the model's ridge regression weights for the instance features vs. dense vectors (Figure~\ref{fig:instance_vs_dense}). As expected, lower brain regions (V1-V3) were more informative about the dense vector, while higher brain regions (V4, LOC, FFA, PPA) carried more information about instance features.

\begin{figure}[h!]
    \centering
    \includegraphics[width=0.45\textwidth]{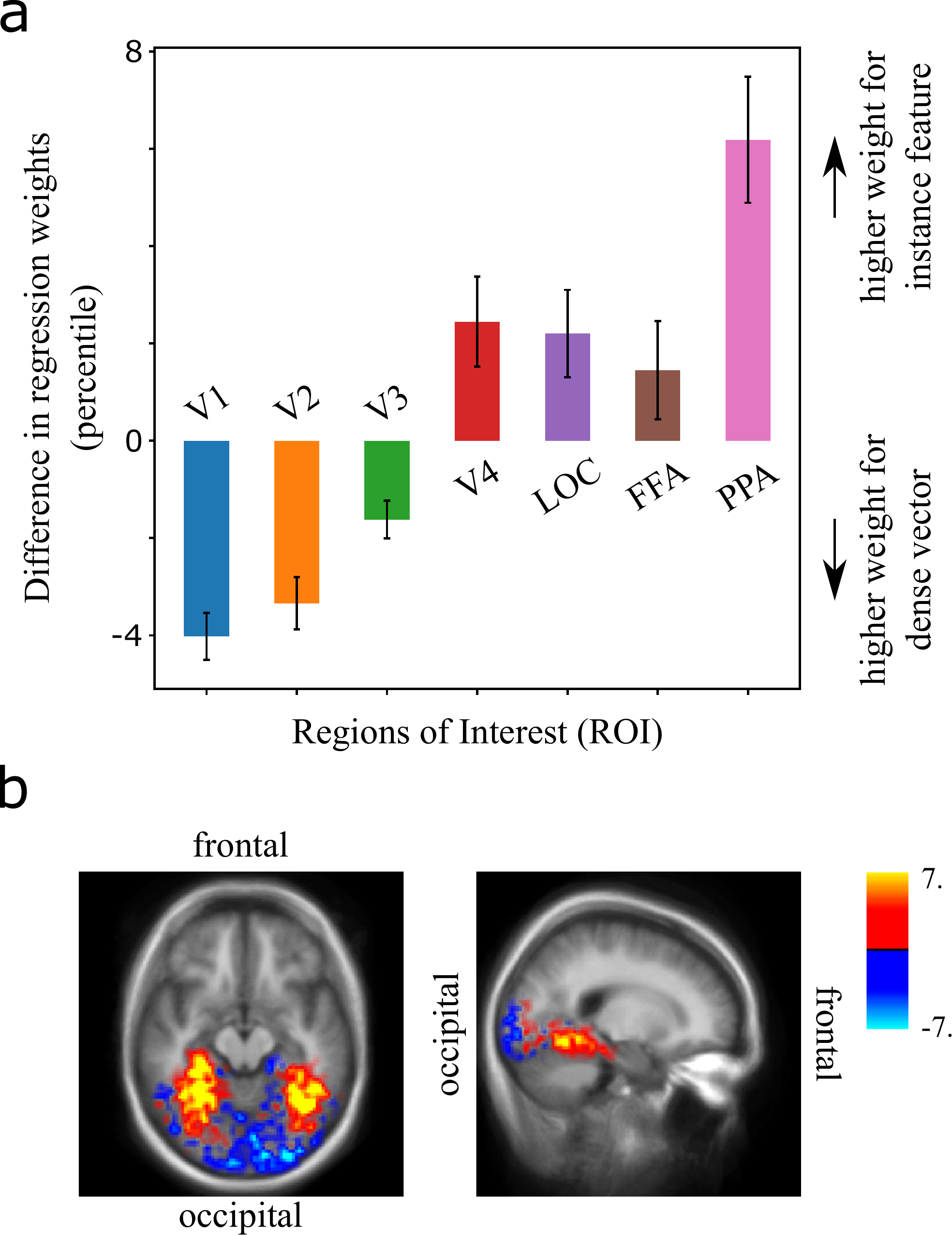}
    \caption{Mapping of instance features vs. dense vectors over brain regions. (a) Difference between the percentiles of the regression weights ($L_1$ norm) for the instance features vs. the dense vector, averaged over voxels in each ROI. Positive values indicate relatively higher weight for instance features compared to the dense vector, and vice versa. Error bars represent standard error of the mean across 5 subjects. (b) Voxel-by-voxel maps (left: axial; right: sagittal) of the difference between the percentiles of the regression weights ($L_1$ norm) for the instance features (red) vs. the dense vector (blue), averaged over the 5 subjects. Dense vector weights are higher in early visual cortex (occipital regions), while instance feature weights are larger in higher visual cortex (temporal regions).}
    \label{fig:instance_vs_dense}
\end{figure}

\begin{figure}
    \centering
    \includegraphics[width=0.5\textwidth]{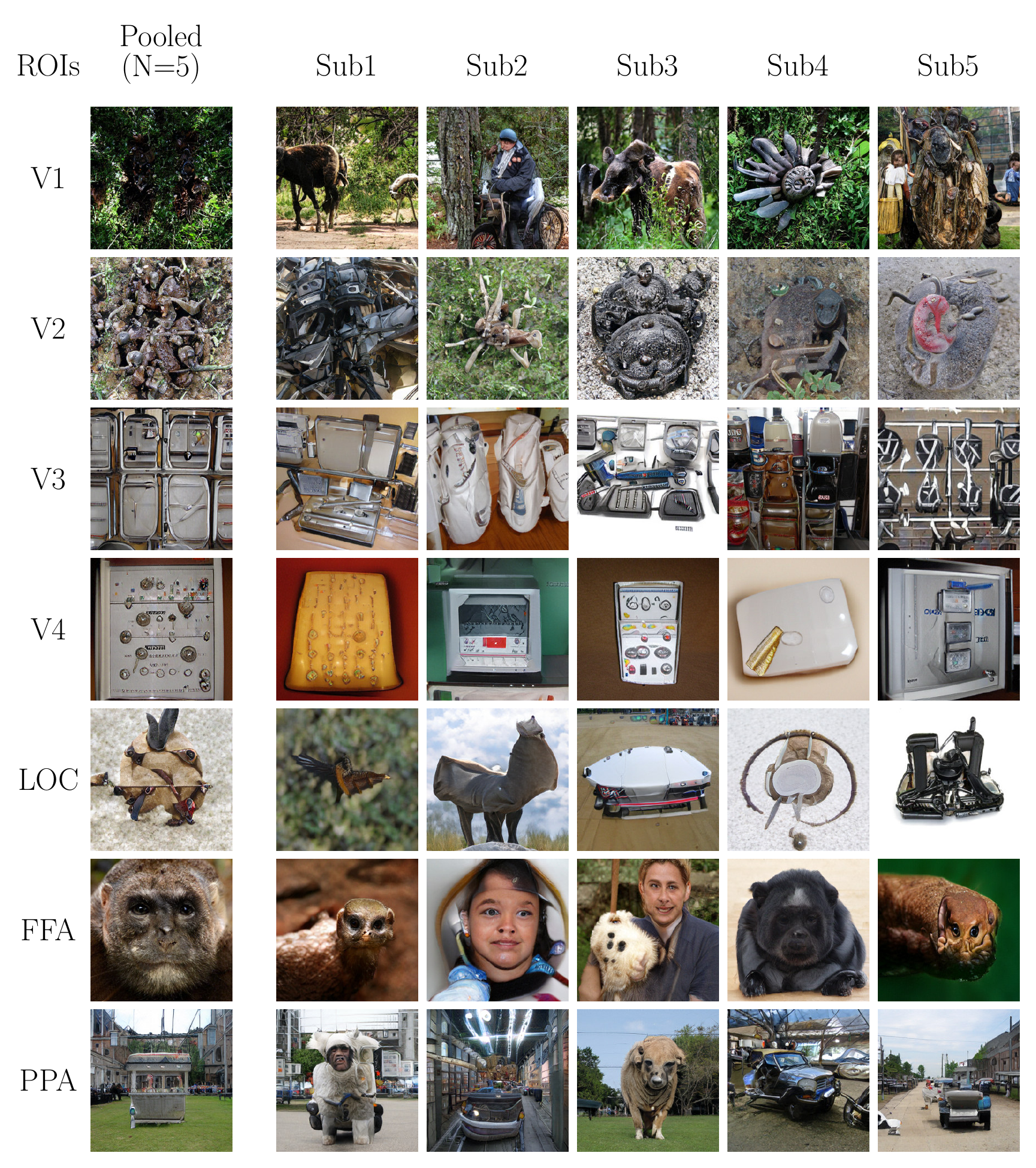}
    \caption{Generated images from synthetic fMRI patterns constructed by activating all voxels in a specific brain region-of-interest (ROI), and none outside of the ROI. The rows represent various brain regions: V1, V2, V3, V4, LOC, FFA, and PPA. The first column is generated after averaging the brain-predicted latent variables for all five subjects. The following columns are for individual subjects.}
    \label{fig:roi_maximization}
\end{figure}

Next, we use our brain decoding model to visualize the ``optimal'' stimulus for each brain region. Instead of using fMRI patterns recorded from subjects, we synthesized seven patterns, one for each ROI, with a value of 1 for all voxels inside the ROI and 0 outside. We provided these synthetic patterns to the three trained ridge regression models to obtain predicted latent variables (as described in Figure~\ref{fig:decoding_reconstruction}). To mitigate the scaling problem, we normalized instance features to have unit norms. We then passed the predicted latent variables through the IC-GAN generator to generate images. 

Previously, Gu et al.~\cite{gu2022neurogen} synthesized optimal images for different ROIs using a BigGAN generator and a feature extractor. They iteratively optimized the latent variables of the generator  in such a way that predicted fMRI patterns (obtained via the feature extractor) maximized activation in a specific ROI. In contrast, our method involves a single pass through our image reconstruction pipeline, and does not require iterative optimization of the latent variables. Figure~\ref{fig:roi_maximization} presents the generated images from each subject (second to sixth columns), together with reconstructions using averaged latent variables across all five subjects (first column). 

In lower visual cortex (V1-V2), basic textures (foliage, trees, stones) are produced rather than (or in addition to) identifiable objects. The textures emphasize the periphery of the visual field, in line with the fact that V1-V2 have small receptive fields that can be positioned at high visual eccentricity. For V3 and V4, the generated textures present more regularity than V1 and V2, and we begin to see visuals close to objects with multiple parts, including text-like symbols, notably in V4. LOC is known for its selectivity to object shapes; when maximizing this region's response, IC-GAN generates complete objects at the center of the image, rather than extended textures. At this stage, the visual periphery appears empty or blurry, in contrast with the crisp peripheral textures produced for V1-V2. In FFA, a high-level region known for its selectivity to face images, IC-GAN generates human and animal faces. The presence of animal faces is not unexpected, since the ImageNet dataset (on which IC-GAN was trained) contains many more animal images than human images. Some previous experimental and computational work~\cite{gu2022neurogen,blonder2004regional} also suggests that fusiform regions may show a preferential response to animals, and particularly dogs. Nonetheless, the model still generates human face images for two of the subjects. The last ROI from the higher visual cortex is PPA, known for its selectivity to environmental scenes like indoor and outdoor places. IC-GAN also generates indoor and outdoor places when the voxels of this region are activated. Some of the images have an object in the center of the scene; this might be caused by the training of the IC-GAN model on ImageNet--an object-centered dataset. It is worth noting that PPA-optimized images produce more details in the visual periphery than FFA-optimized images; this is compatible with the known difference in preferential eccentricity between the two regions~\cite{levy2001center}. Overall, the outcomes of this analysis are consistent with findings from the neuroscience literature, indicating that our IC-GAN-based model learned to appropriately decode visual feature selectivity in the brain. Most importantly, the method allows us to \emph{directly visualize} this selectivity, rather than inferring it from extended experiments.

\section{Discussion}
In this paper, we presented a framework for natural image reconstruction from fMRI patterns using the IC-GAN model, pretrained on ImageNet. First, we extracted instance features, noise vectors, and dense vectors from training images, and trained ridge regression models from fMRI patterns to these latent variables. With these regression models, we decoded latent variables from the test fMRI patterns, and finally reconstructed images with the IC-GAN generator.

Many previous studies implemented fMRI reconstruction frameworks with deep generative models. However, these models were able to reconstruct either low-level or high-level features of the images. Our method demonstrated state-of-the-art performance on reconstructing semantic (high-level) attributes of the images, both qualitatively and quantitatively, while generating naturalistic-looking images. Meanwhile, compared to other semantically oriented models (e.g. Mozafari et al.~\cite{mozafari2020reconstructing}, an approach based on BigBiGAN), it was able to maintain more low-level details. Furthermore, we could use our fMRI-based image reconstruction model to visualize images decoded from synthetic fMRI patterns, designed to maximize activations in specific brain ROIs. The results of this analysis were aligned with the existing neuroscience literature, opening a range of possibilities for future brain exploration and visualization techniques.

We acknowledge that there is still room for improving our model, especially in terms of better reproducing low-level details. This may be achieved in future work by improving our optimization of the noise and dense vectors, or by pairing IC-GAN with other generative networks more focused on low-level image properties.

% \section*{References}
\bibliographystyle{IEEEtran}

\bibliography{refs}

\begin{thebibliography}{10}
\providecommand{\url}[1]{#1}
\csname url@samestyle\endcsname
\providecommand{\newblock}{\relax}
\providecommand{\bibinfo}[2]{#2}
\providecommand{\BIBentrySTDinterwordspacing}{\spaceskip=0pt\relax}
\providecommand{\BIBentryALTinterwordstretchfactor}{4}
\providecommand{\BIBentryALTinterwordspacing}{\spaceskip=\fontdimen2\font plus
\BIBentryALTinterwordstretchfactor\fontdimen3\font minus
  \fontdimen4\font\relax}
\providecommand{\BIBforeignlanguage}[2]{{%
\expandafter\ifx\csname l@#1\endcsname\relax
\typeout{** WARNING: IEEEtran.bst: No hyphenation pattern has been}%
\typeout{** loaded for the language `#1'. Using the pattern for}%
\typeout{** the default language instead.}%
\else
\language=\csname l@#1\endcsname
\fi
#2}}
\providecommand{\BIBdecl}{\relax}
\BIBdecl

\bibitem{thirion2006inverse}
B.~Thirion, E.~Duchesnay, E.~Hubbard, J.~Dubois, J.-B. Poline, D.~Lebihan, and
  S.~Dehaene, ``Inverse retinotopy: inferring the visual content of images from
  brain activation patterns,'' \emph{Neuroimage}, vol.~33, no.~4, pp.
  1104--1116, 2006.

\bibitem{kamitani2005decoding}
Y.~Kamitani and F.~Tong, ``Decoding the visual and subjective contents of the
  human brain,'' \emph{Nature neuroscience}, vol.~8, no.~5, pp. 679--685, 2005.

\bibitem{haynes2005predicting}
J.-D. Haynes and G.~Rees, ``Predicting the orientation of invisible stimuli
  from activity in human primary visual cortex,'' \emph{Nature neuroscience},
  vol.~8, no.~5, pp. 686--691, 2005.

\bibitem{haxby2001distributed}
J.~V. Haxby, M.~I. Gobbini, M.~L. Furey, A.~Ishai, J.~L. Schouten, and
  P.~Pietrini, ``Distributed and overlapping representations of faces and
  objects in ventral temporal cortex,'' \emph{Science}, vol. 293, no. 5539, pp.
  2425--2430, 2001.

\bibitem{cox2003functional}
D.~D. Cox and R.~L. Savoy, ``Functional magnetic resonance imaging
  (fmri)“brain reading”: detecting and classifying distributed patterns of
  fmri activity in human visual cortex,'' \emph{Neuroimage}, vol.~19, no.~2,
  pp. 261--270, 2003.

\bibitem{kay2008identifying}
K.~N. Kay, T.~Naselaris, R.~J. Prenger, and J.~L. Gallant, ``Identifying
  natural images from human brain activity,'' \emph{Nature}, vol. 452, no.
  7185, pp. 352--355, 2008.

\bibitem{miyawaki2008visual}
Y.~Miyawaki, H.~Uchida, O.~Yamashita, M.-a. Sato, Y.~Morito, H.~C. Tanabe,
  N.~Sadato, and Y.~Kamitani, ``Visual image reconstruction from human brain
  activity using a combination of multiscale local image decoders,''
  \emph{Neuron}, vol.~60, no.~5, pp. 915--929, 2008.

\bibitem{schoenmakers2013linear}
S.~Schoenmakers, M.~Barth, T.~Heskes, and M.~Van~Gerven, ``Linear
  reconstruction of perceived images from human brain activity,''
  \emph{NeuroImage}, vol.~83, pp. 951--961, 2013.

\bibitem{vanrullen2019reconstructing}
R.~VanRullen and L.~Reddy, ``Reconstructing faces from fmri patterns using deep
  generative neural networks,'' \emph{Communications biology}, vol.~2, no.~1,
  pp. 1--10, 2019.

\bibitem{shen2019deep}
G.~Shen, T.~Horikawa, K.~Majima, and Y.~Kamitani, ``Deep image reconstruction
  from human brain activity,'' \emph{PLoS computational biology}, vol.~15,
  no.~1, p. e1006633, 2019.

\bibitem{casanova2021instanceconditioned}
\BIBentryALTinterwordspacing
A.~Casanova, M.~Careil, J.~Verbeek, M.~Drozdzal, and A.~Romero,
  ``Instance-conditioned {GAN},'' in \emph{Advances in Neural Information
  Processing Systems}, A.~Beygelzimer, Y.~Dauphin, P.~Liang, and J.~W. Vaughan,
  Eds., 2021. [Online]. Available:
  \url{https://openreview.net/forum?id=aUuTEEcyY_}
\BIBentrySTDinterwordspacing

\bibitem{jaiswal2021survey}
A.~Jaiswal, A.~R. Babu, M.~Z. Zadeh, D.~Banerjee, and F.~Makedon, ``A survey on
  contrastive self-supervised learning,'' \emph{Technologies}, vol.~9, no.~1,
  p.~2, 2021.

\bibitem{pividori2019exploiting}
M.~Pividori, G.~L. Grinblat, and L.~C. Uzal, ``Exploiting gan internal capacity
  for high-quality reconstruction of natural images,'' \emph{arXiv preprint
  arXiv:1911.05630}, 2019.

\bibitem{seeliger2018generative}
\BIBentryALTinterwordspacing
K.~Seeliger, U.~Güçlü, L.~Ambrogioni, Y.~Güçlütürk, and M.~{van Gerven},
  ``Generative adversarial networks for reconstructing natural images from
  brain activity,'' \emph{NeuroImage}, vol. 181, pp. 775--785, 2018. [Online].
  Available:
  \url{https://www.sciencedirect.com/science/article/pii/S105381191830658X}
\BIBentrySTDinterwordspacing

\bibitem{beliy2019voxels}
R.~Beliy, G.~Gaziv, A.~Hoogi, F.~Strappini, T.~Golan, and M.~Irani, ``From
  voxels to pixels and back: Self-supervision in natural-image reconstruction
  from fmri,'' \emph{Advances in Neural Information Processing Systems},
  vol.~32, 2019.

\bibitem{gaziv2020self}
G.~Gaziv, R.~Beliy, N.~Granot, A.~Hoogi, F.~Strappini, T.~Golan, and M.~Irani,
  ``Self-supervised natural image reconstruction and rich semantic
  classification from brain activity,'' \emph{bioRxiv}, 2020.

\bibitem{mozafari2020reconstructing}
M.~Mozafari, L.~Reddy, and R.~VanRullen, ``Reconstructing natural scenes from
  fmri patterns using bigbigan,'' in \emph{2020 International joint conference
  on neural networks (IJCNN)}.\hskip 1em plus 0.5em minus 0.4em\relax IEEE,
  2020, pp. 1--8.

\bibitem{donahue2019large}
J.~Donahue and K.~Simonyan, ``Large scale adversarial representation
  learning,'' \emph{Advances in Neural Information Processing Systems},
  vol.~32, 2019.

\bibitem{ren2021reconstructing}
Z.~Ren, J.~Li, X.~Xue, X.~Li, F.~Yang, Z.~Jiao, and X.~Gao, ``Reconstructing
  seen image from brain activity by visually-guided cognitive representation
  and adversarial learning,'' \emph{NeuroImage}, vol. 228, p. 117602, 2021.

\bibitem{deng2009imagenet}
J.~Deng, W.~Dong, R.~Socher, L.-J. Li, K.~Li, and L.~Fei-Fei, ``Imagenet: A
  large-scale hierarchical image database,'' in \emph{2009 IEEE conference on
  computer vision and pattern recognition}.\hskip 1em plus 0.5em minus
  0.4em\relax Ieee, 2009, pp. 248--255.

\bibitem{karras2020analyzing}
T.~Karras, S.~Laine, M.~Aittala, J.~Hellsten, J.~Lehtinen, and T.~Aila,
  ``Analyzing and improving the image quality of stylegan,'' in
  \emph{Proceedings of the IEEE/CVF conference on computer vision and pattern
  recognition}, 2020, pp. 8110--8119.

\bibitem{brock2018large}
\BIBentryALTinterwordspacing
A.~Brock, J.~Donahue, and K.~Simonyan, ``Large scale {GAN} training for high
  fidelity natural image synthesis,'' in \emph{International Conference on
  Learning Representations}, 2019. [Online]. Available:
  \url{https://openreview.net/forum?id=B1xsqj09Fm}
\BIBentrySTDinterwordspacing

\bibitem{mirza2014conditional}
M.~Mirza and S.~Osindero, ``Conditional generative adversarial nets,''
  \emph{arXiv preprint arXiv:1411.1784}, 2014.

\bibitem{caron2020unsupervised}
M.~Caron, I.~Misra, J.~Mairal, P.~Goyal, P.~Bojanowski, and A.~Joulin,
  ``Unsupervised learning of visual features by contrasting cluster
  assignments,'' \emph{Advances in Neural Information Processing Systems},
  vol.~33, pp. 9912--9924, 2020.

\bibitem{he2016deep}
K.~He, X.~Zhang, S.~Ren, and J.~Sun, ``Deep residual learning for image
  recognition,'' in \emph{Proceedings of the IEEE conference on computer vision
  and pattern recognition}, 2016, pp. 770--778.

\bibitem{konkle2022self}
T.~Konkle and G.~A. Alvarez, ``A self-supervised domain-general learning
  framework for human ventral stream representation,'' \emph{Nature
  Communications}, vol.~13, no.~1, pp. 1--12, 2022.

\bibitem{zhuang2021unsupervised}
C.~Zhuang, S.~Yan, A.~Nayebi, M.~Schrimpf, M.~C. Frank, J.~J. DiCarlo, and
  D.~L. Yamins, ``Unsupervised neural network models of the ventral visual
  stream,'' \emph{Proceedings of the National Academy of Sciences}, vol. 118,
  no.~3, 2021.

\bibitem{hansen2001completely}
N.~Hansen and A.~Ostermeier, ``Completely derandomized self-adaptation in
  evolution strategies,'' \emph{Evolutionary Computation}, vol.~9, no.~2, pp.
  159--195, 2001.

\bibitem{zhang2018unreasonable}
R.~Zhang, P.~Isola, A.~A. Efros, E.~Shechtman, and O.~Wang, ``The unreasonable
  effectiveness of deep features as a perceptual metric,'' in \emph{Proceedings
  of the IEEE conference on computer vision and pattern recognition}, 2018, pp.
  586--595.

\bibitem{simonyan2014very}
K.~Simonyan and A.~Zisserman, ``Very deep convolutional networks for
  large-scale image recognition,'' \emph{arXiv preprint arXiv:1409.1556}, 2014.

\bibitem{horikawa2017generic}
T.~Horikawa and Y.~Kamitani, ``Generic decoding of seen and imagined objects
  using hierarchical visual features,'' \emph{Nature communications}, vol.~8,
  no.~1, pp. 1--15, 2017.

\bibitem{wang2004image}
Z.~Wang, A.~C. Bovik, H.~R. Sheikh, and E.~P. Simoncelli, ``Image quality
  assessment: from error visibility to structural similarity,'' \emph{IEEE
  transactions on image processing}, vol.~13, no.~4, pp. 600--612, 2004.

\bibitem{szegedy2016rethinking}
C.~Szegedy, V.~Vanhoucke, S.~Ioffe, J.~Shlens, and Z.~Wojna, ``Rethinking the
  inception architecture for computer vision,'' in \emph{Proceedings of the
  IEEE conference on computer vision and pattern recognition}, 2016, pp.
  2818--2826.

\bibitem{radford2021learning}
A.~Radford, J.~W. Kim, C.~Hallacy, A.~Ramesh, G.~Goh, S.~Agarwal, G.~Sastry,
  A.~Askell, P.~Mishkin, J.~Clark \emph{et~al.}, ``Learning transferable visual
  models from natural language supervision,'' in \emph{International Conference
  on Machine Learning}.\hskip 1em plus 0.5em minus 0.4em\relax PMLR, 2021, pp.
  8748--8763.

\bibitem{gu2022neurogen}
Z.~Gu, K.~W. Jamison, M.~Khosla, E.~J. Allen, Y.~Wu, T.~Naselaris, K.~Kay,
  M.~R. Sabuncu, and A.~Kuceyeski, ``Neurogen: activation optimized image
  synthesis for discovery neuroscience,'' \emph{NeuroImage}, vol. 247, p.
  118812, 2022.

\bibitem{blonder2004regional}
L.~X. Blonder, C.~D. Smith, C.~E. Davis, M.~L. Kesler, T.~F. Garrity, M.~J.
  Avison, A.~H. Andersen \emph{et~al.}, ``Regional brain response to faces of
  humans and dogs,'' \emph{Cognitive Brain Research}, vol.~20, no.~3, pp.
  384--394, 2004.

\bibitem{levy2001center}
I.~Levy, U.~Hasson, G.~Avidan, T.~Hendler, and R.~Malach, ``Center--periphery
  organization of human object areas,'' \emph{Nature neuroscience}, vol.~4,
  no.~5, pp. 533--539, 2001.

\end{thebibliography}

\vspace{12pt}

\end{document}